\documentclass[11pt, a4paper]{article}
\usepackage[hyperref]{styles/acl2018}
\pdfoutput=1
\usepackage{times}
\usepackage{latexsym}
\usepackage{amsmath}
\usepackage{styles/multirow}
\usepackage{url}
\usepackage{graphicx}
\usepackage{amsfonts}
\usepackage{amssymb}
\usepackage{fixltx2e}
\usepackage{float}
\usepackage{xcolor}
\usepackage{subfigure}
\usepackage{placeins}

\newcommand{\refeqn}[1]{Equation \ref{#1}}
\newcommand{\reffig}[1]{Figure \ref{#1}}
\newcommand{\reftbl}[1]{Table \ref{#1}}
\newcommand{\refsec}[1]{Section \ref{#1}}

\newcommand{\problem}{HRSI}
\newcommand{\problemfull}{Higher-order Relation Schema Induction}

\newcommand{\system}{TFBA}
\newcommand{\systemfull}{Tensor Factorization with Back-off and Aggregation}

\newcommand{\HCB}{HardClust}

\newcommand{\nary}{\texttt{n}-ary }

\newcommand{\tuples}{\mathbb{T}}

\newcommand\norm[1]{\left\lVert#1\right\rVert}

\DeclareMathAlphabet\ten{OMS}{cmsy}{b}{n} %%usage: \mathbfcal{W}
\def\mat#1{\mbox{\bf #1}}%% usage: \mat{W}.

\title{Higher-order Relation Schema Induction using Tensor Factorization with Back-off and Aggregation}

\author{Madhav Nimishakavi \\
  Indian Institute of Science \\
  Bangalore \\
  {\tt madhav@iisc.ac.in} \\\And
  Manish Gupta \\
  Microsoft \\
  Hyderabad \\
  {\tt manishg@microsoft.com} \\\And
  Partha Talukdar \\
  Indian Institute of Science\\
  Bangalore \\
  {\tt ppt@iisc.ac.in}\\
  }

\date{}

\aclfinalcopy % Uncomment this line for the final submission
\begin{document}
\maketitle

\begin{abstract}
Relation Schema Induction (RSI) is the problem of identifying  type signatures of arguments of relations from unlabeled text.  Most of the previous work in this area have focused only on   \emph{binary} RSI, i.e., inducing only the 
subject and object type signatures per relation. However, in practice, many relations are \emph{high-order}, i.e., they have more than two arguments and inducing type signatures of all arguments is necessary. For example, in the sports domain, inducing a schema \textit{win(WinningPlayer, OpponentPlayer, Tournament, Location)} is more informative than inducing just \textit{win(WinningPlayer, OpponentPlayer)}. We refer to this problem as \problemfull{} (\problem{}). In this paper, we propose \systemfull{} (\system{}), a novel framework for the \problem{} problem. 
To the best of our knowledge, this is the first attempt at inducing higher-order relation schemata from unlabeled text. Using the experimental analysis on three real world datasets, we show how \system{} helps in dealing with sparsity and induce higher order schemata. 
%We will make all the code and datasets used in the paper publicly available upon publication of the paper.
\end{abstract}

\section{Introduction}\label{sec:intro}

Building Knowledge Graphs (KGs) out of unstructured data is an area of active research. Research in this has resulted in the construction of several large scale KGs, such as NELL \cite{NELL-aaai15}, Google Knowledge Vault \cite{dong2014knowledge} and YAGO \cite{suchanek2007yago}. These KGs consist of millions of entities and beliefs involving those entities. 
Such KG construction methods are schema-guided as they require the list of input relations and their schemata (e.g., \textit{playerPlaysSport(Player, Sport)}). In other words, knowledge of schemata is an important first step towards building such KGs.

While beliefs in such KGs are usually binary (i.e., involving two entities), many beliefs of interest go beyond two entities. For example, in the sports domain, one may be interested in beliefs of the form \textit{win(Roger Federer, Nadal, Wimbledon, London)}, which is an instance of the high-order (or n-ary) relation \textit{win} whose schema is given by \textit{win(WinningPlayer, OpponentPlayer, Tournament, Location)}. 
We refer to the problem of inducing such relation schemata involving multiple arguments as \problemfull{} (\problem{}). In spite of its importance, \problem{} is mostly unexplored.

Recently, tensor factorization-based methods have been proposed for \emph{binary} relation schema induction \cite{nimishakavi-saini-talukdar:2016:EMNLP2016}, with gains in both speed and accuracy over previously proposed generative models. To the best of our  knowledge, tensor factorization methods have not been used for \problem{}. %inducing higher order schemata for relations. 
We address this gap in this paper.

Due to data sparsity, straightforward adaptation of tensor factorization from \cite{nimishakavi-saini-talukdar:2016:EMNLP2016} to \problem{} %event schema induction 
is not feasible, as we shall see in Section~\ref{sec:failed_attempt}. We overcome this challenge in this paper, and make the following contributions.

\begin{itemize}
	\item We propose \systemfull{} (\system{}), a novel tensor factorization-based method for Higher-order RSI (\problem{}). In order to overcome data sparsity, \system{} \emph{backs-off} and jointly factorizes multiple lower-order tensors derived from an extremely sparse higher-order tensor.
	\item As an aggregation step, we propose a constrained clique mining step which constructs the higher-order schemata from multiple binary schemata. 
	\item Through experiments on multiple real-world datasets, we show the effectiveness of \system{} for \problem{}.
\end{itemize} 

Source code of \system{} is available at \url{https://github.com/madhavcsa/TFBA}.

The remainder of the paper is organized as follows. We discuss related work in Section~\ref{sec:related}. In Section~\ref{sec:failed_attempt}, we first motivate why a back-off strategy is needed for \problem{}, rather than factorizing the higher-order tensor. Further, we discuss the proposed \system{} framework in Section~\ref{sec:model}. In Section~\ref{sec:exp}, we demonstrate the effectiveness of the proposed approach using multiple real world datasets. We conclude with a brief summary in Section~\ref{sec:conclusion}.

\section{Related Work}
\label{sec:related}

In this section, we discuss related works in two broad areas: schema induction, and tensor and matrix factorizations. 

\textbf{Schema Induction:} Most work on inducing schemata for relations has been in the binary setting   \cite{Mohamed11discoveringrelations,kblda:movshovitzattias-wcohen:2015:ACL,nimishakavi-saini-talukdar:2016:EMNLP2016}. 
 \citet{McDonald:2005:SAC:1219840.1219901} and \citet{Peng2017} extract \nary{} relations from Biomedical documents, but do not induce the schema, i.e., type signature of the \nary{} relations. There has been significant amount of work on 
 Semantic Role Labeling \cite{lang-lapata:2011:ACL-HLT2011,titov-khoddam:2015:NAACL-HLT,RothLapata:16}, which can be considered as \nary{} relation extraction. However, we are interested in inducing the 
 schemata, i.e., the type signature of these relations.
 Event Schema Induction  is the problem of inducing schemata for events in the corpus \cite{balasubramanian-EtAl:2013:EMNLP,Chambers13,nguyen-EtAl:2015:ACL-IJCNLP1}. Recently, a model for event representations is proposed in \cite{Weber2018}.
 \begin{table*}[t]
 \centering
 \small
  \begin{tabular}{|l|l|}
  \hline
   Notation & Definition \\
   \hline
	\hline
   $\mathbb{R}_{+}$ & Set of non-negative reals. \\
   \hline
   $\ten{X} \in \mathbb{R}_{+}^{n_1 \times n_2 \times \ldots \times  n_N}$ & $N^{\mathrm{th}}$ -order non-negative tensor. \\
   \hline
   $\ten{X}_{(i)}$ & mode-$i$ matricization of tensor $\ten{X}$ . Please see \cite{kolda2009tensor} for details.\\
   \hline
   $\mat{A} \in \mathbb{R}_{+}^{n \times r}$ & Non-negative matrix of order $n \times r$. \\
   \hline
   $*$ & Hadamard product: $(\mat{A} * \mat{B})_{i,j} = \mat{A}_{i,j} \times \mat{B}_{i,j}.$\\   
   \hline
  \end{tabular}
\caption{\label{tbl:notations} Notations used in the paper.}
 \end{table*}

  \citet{cheung-poon-vanderwende:2013:NAACL-HLT} propose a probabilistic model for inducing frames from text. Their notion of frame is closer to that of scripts \cite{schank:77a}. 
 Script learning is the process of automatically inferring sequence of events from text \cite{Mooney:1985:LSN:1625135.1625267}. There is a fair amount of recent work in statistical script learning \cite{Pichotta:2016:LSS:3016100.3016293}, \cite{pichotta:eacl14}. While script learning deals with the sequence of events, we try to find the schemata of relations at a corpus level. %, which can be treated as predicate centric event schema induction.
  \citet{Ferraro:2016:UBM:3016100.3016265} propose a unified Bayesian model for scripts, frames and events. 
 Their model tries to capture all levels of Minsky Frame structure \cite{Minsky:1974:FRK:889222}, however we work with the surface semantic frames.

 \textbf{Tensor and Matrix Factorizations: } Matrix factorization and joint tensor-matrix factorizations have been used for the problem of predicting links in the Universal Schema setting  \cite{RiedelYMM13,singh2015towards}.
 \citet{ChenWGR15}  use matrix factorizations for the problem of finding semantic slots for unsupervised spoken language understanding. Tensor factorization methods are also used in factorizing knowledge graphs \cite{export:226677,Nickel:2012:FYS:2187836.2187874}. Joint matrix and tensor factorization frameworks, where the matrix provides additional information, is proposed in \cite{Acar2013} and \cite{conf/kdd/WangCGDKCMS15}. These models are based
 on PARAFAC \cite{harshman1970fpp}, a tensor factorization model which approximates the given tensor as a sum of rank-1 tensors. A boolean Tucker decomposition for discovering facts is proposed in \cite{Erdos:2013:DFB:2505515.2507846}. In this paper, we use a modified version (Tucker2) of Tucker decomposition \cite{Tuck1963a}.
  
 RESCAL \cite{Nickel_athree-way} is a simplified Tucker model suitable for relational learning.
 Recently, SICTF \cite{nimishakavi-saini-talukdar:2016:EMNLP2016}, a variant of RESCAL with side information, is used for the problem of schema induction for binary relations. 
 SICTF cannot be directly used to induce higher order schemata, as the higher-order tensors involved in inducing such schemata tend to be extremely sparse. 
 \system{} overcomes these challenges to induce higher-order relation schemata by performing Non-Negative Tucker-style factorization of sparse tensor while utilizing a back-off strategy, as explained in the next section.

\section{Higher Order Relation Schema Induction using Back-off Factorization} \label{sec:method}

\begin{figure*}[t]

\centering
\includegraphics[scale=0.35]{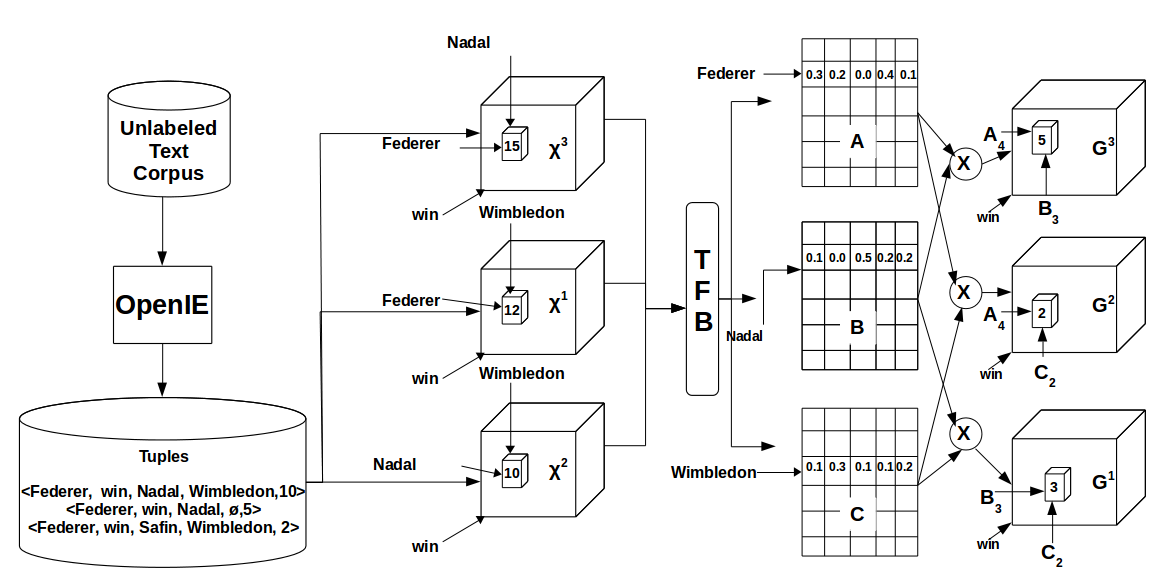}
\caption{\label{fig:arch} {\bf Overview of Step 1 of \system{}}. Rather than factorizing the higher-order tensor $\ten{X}$, \system{} performs joint Tucker decomposition of multiple 3-mode tensors, $\ten{X}^1$, $\ten{X}^2$, and $\ten{X}^3$, derived out of $\ten{X}$. This joint factorization is performed using shared latent factors $\mat{A}$,  $\mat{B}$, and $\mat{C}$. This results in binary schemata, each of which is stored as a cell in one of the core tensors $\mathcal{G}^1$, $\mathcal{G}^2$, and $\mathcal{G}^3$. Please see \refsec{sec:backoff_factor} for details.}
\end{figure*}

\begin{figure}[t]
\centering
\includegraphics[scale=0.33]{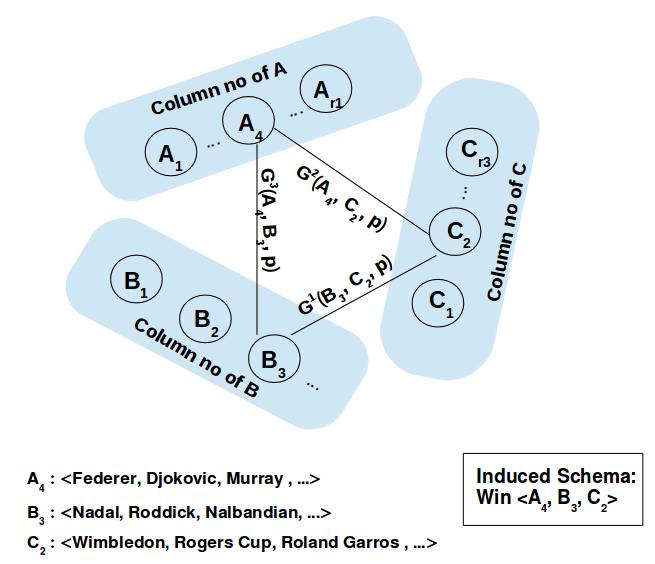}
\caption{\label{fig:schemata}{\bf Overview of Step 2 of \system{}}. Induction of  higher-order schemata from the tri-partite graph formed from the columns of matrices $A$, $B$, and $C$. Triangles in this graph (solid) represent a 3-ary  schema, n-ary schemata for n $> 3$ can 
be induced from the 3-ary schemata. Please refer to \refsec{sec:schema} for  details.} 
\end{figure}

In this section, we start by discussing the approach of factorizing a higher-order tensor and provide the motivation for back-off strategy. 
Next,  we discuss the proposed \system{} approach in detail. Please refer to \reftbl{tbl:notations} for notations used in this paper.

\subsection{Factorizing a Higher-order Tensor}\label{sec:failed_attempt}
	 Given a text corpus, we use OpenIEv5 \cite{mausam2016} to extract tuples. 
	Consider the following sentence ``\textit{Federer won against  Nadal at Wimbledon.}''. Given this sentence, OpenIE extracts the 4-tuple \textit{(Federer, won, against Nadal, at Wimbledon)}. We lemmatize the relations in the tuples and only consider the noun phrases as arguments. Let $\mathbb{T}$ represent the set of these 4-tuples.	
	We can construct a 4-order tensor $\ten{X} \in \mathbb{R}_{+}^{n_1 \times n_2 \times n_3 \times m}$ from $\tuples{}$. 
	Here, $n_1$ is the number of {\it subject} noun phrases (NPs), $n_2$ is the number of {\it object} NPs, $n_3$ is the number of {\it other} NPs, and $m$ is the number of relations in $\tuples{}$. Values in the tensor correspond to the frequency of the tuples. 
	In case of 5-tuples of the form {\it (subject, relation, object, other-1, other-2)}, we split the 5-tuples into two 4-tuples of the form {\it (subject, relation, object, other-1)} and {\it (subject, relation, object, other-2)} and frequency of these 4-tuples is considered to be same as the original 5-tuple.
	Factorizing the tensor $\ten{X}$ results in discovering latent categories of NPs, which help in inducing the schemata. We propose the following approach to factorize $\ten{X}$.	
	
	\begin{multline*}
	 \min\limits_{\ten{G}, \mat{A}, \mat{B}, \mat{C}} \norm{ \ten{X} - \ten{G} \times_1 \mat{A} \times_2 \mat{B} \times_3 \mat{C} \times_4 \mat{I} }_F^2 \\
		+ \lambda_a \norm{\mat{A}}_F^2 + \lambda_b \norm{\mat{B}}_F^2 + \lambda_c \norm{\mat{C}}_F^2  ,
	\end{multline*}

	where,

	\begin{multline*}	
	\mat{A} \in \mathbb{R}_{+}^{n_1 \times r_1}, \mat{B} \in \mathbb{R}_{+}^{n_2 \times r_2}, \mat{C} \in \mathbb{R}_{+}^{n_3 \times r_3}, \\
	  \ten{G} \in \mathbb{R}_{+}^{r_1 \times r_2 \times r_3 \times m}, \lambda_a \geq 0, \lambda_b \geq 0 \text{~and~} \lambda_c \geq 0. 
	\end{multline*}
	
	Here, $\mat{I}$ is the identity matrix.
    Non-negative updates for the variables can be obtained following \cite{Lee00algorithmsfor}. Similar to \cite{nimishakavi-saini-talukdar:2016:EMNLP2016}, schemata induced will be of the form \textit{relation} $\langle \mat{A}_i, \mat{B}_j, \mat{C}_k \rangle$. Here, $\mat{P}_i$ represents the $i^{\text{th}}$ column of a matrix $\mat{P}$. 
    $\mat{A}$ is the embedding matrix of subject NPs in $\tuples{}$ (i.e., mode-1 of $\ten{X}$), $r_1$ is the embedding rank in mode-1 which is the number of  latent categories of subject NPs. Similarly, $\mat{B}$ and $\mat{C}$ are the embedding matrices of object NPs and other NPs respectively. $r_2$ and $r_3$ are the number of latent categories of object NPs and other NPs respectively. $\ten{G}$ is the core tensor. $\lambda_a$, $\lambda_b$ and $\lambda_c$ are the regularization weights.

	However, the 4-order tensors are heavily sparse for all  the datasets we consider in this work. The sparsity ratio of this 4-order tensor for all the datasets is of the order 1e-7. As a result of the extreme sparsity, this approach fails to learn any schemata.
	Therefore, we propose a more successful  back-off strategy for higher-order RSI in the next section.

  \subsection{\system{}: Proposed Framework}
  \label{sec:model}
  
  To alleviate the problem of sparsity, we construct three tensors $\ten{X}^3$, $\ten{X}^2$, and $\ten{X}^1$ from $\mathbb{T}$ as follows:
  
  \begin{itemize}
   \item $\ten{X}^3 \in \mathbb{R}_{+}^{n_1 \times n_2 \times m}$ is constructed out of the tuples in $\tuples{}$ by dropping the \textit{other} argument and aggregating resulting tuples, i.e., $\ten{X}^3_{i,j,p} = \sum_{k=1}^{n_3} \ten{X}_{i,j,k,p}$. For example, 4-tuples $\langle${\it(Federer, Win, Nadal, Wimbledon), 10}$\rangle$ and $\langle${\it(Federer, Win, Nadal, Australian Open), 5}$\rangle$ will be aggregated to form a triple $\langle${\it(Federer, Win, Nadal), 15}$\rangle$.
   \item $\ten{X}^2  \in \mathbb{R}_{+}^{n_1 \times n_3 \times m}$ is constructed out of the tuples in $\tuples{}$ by dropping the \textit{object} argument and aggregating resulting tuples  i.e.,  $\ten{X}^2_{i,j,p} = \sum_{k=1}^{n_2} \ten{X}_{i,k,j,p}$.
   \item $\ten{X}^1  \in \mathbb{R}_{+}^{n_2 \times n_3 \times m}$ constructed out of the tuples in $\tuples{}$ by dropping the \textit{subject} argument and aggregating resulting tuples i.e., $\ten{X}^1_{i,j,p} = \sum_{k=1}^{n_1} \ten{X}_{k,i,j,p}$.
  \end{itemize}

  The proposed framework \system{} for inducing higher order schemata involves the following two steps.
  \begin{itemize}
  	\item {\bf Step 1}: In this step, \system{} factorizes multiple lower-order overlapping tensors, $\ten{X}^1$, $\ten{X}^2$, and $\ten{X}^3$, derived from $\ten{X}$ to induce binary schemata. This step is illustrated in  \reffig{fig:arch} and we discuss details in \refsec{sec:backoff_factor}.
  	\item {\bf Step 2}: In this step, \system{} connects multiple binary schemata identified above to induce higher-order schemata. The method accomplishes this by solving a constrained clique problem. This step is illustrated in  \reffig{fig:schemata}   and we discuss the details in \refsec{sec:schema}.
  \end{itemize}
  
\subsubsection{Step 1: Back-off Tensor Factorization}
\label{sec:backoff_factor}

	A schematic overview of this step is shown in \reffig{fig:arch}. \system{} first preprocesses the corpus and extracts OpenIE tuple set $\tuples{}$ out of it. The 4-mode tensor $\ten{X}$ is constructed out of $\tuples{}$. Instead of performing factorization of the higher-order tensor $\ten{X}$ as in \refsec{sec:failed_attempt}, \system{} creates three tensors out of $\ten{X}$:  $\ten{X}^1_{n_2\times n_3 \times m}, \ten{X}^2_{n_1\times n_3 \times m}$ and $\ten{X}^3_{n_1\times n_2 \times m}$.  
	
  \system{} performs a coupled non-negative Tucker factorization of the input tensors $\ten{X}^1, \ten{X}^2$ and $\ten{X}^3$ by solving the following optimization problem.
  \begin{multline}
  \label{eqn:jtf_eqn}
   \underset{\ten{G}^1, \ten{G}^2, \ten{G}^3}{\min\limits_{\mat{A},\mat{B},\mat{C}}} f(\ten{X}^3, \ten{G}^3, \mat{A}, \mat{B}) + f(\ten{X}^2, \ten{G}^2, \mat{A}, \mat{C}) \\
   \qquad \qquad \qquad+ f(\ten{X}^1, \ten{G}^1, \mat{B}, \mat{C}) \\
   + \lambda_a \norm{\mat{A}}_F ^2 + \lambda_b \norm{\mat{B}}_F ^2 +\lambda_c \norm{\mat{C}}_F ^2  ,   
  \end{multline}
where, 
\begin{multline*}
 f(\ten{X}^i, \ten{G}^i, \mat{P}, \mat{Q}) =  \norm{\ten{X}^{i} - \ten{G}^{i} \times_1 \mat{P} \times_2 \mat{Q} \times_3 \mat{I}}_F ^2 \\
 \mat{A} \in 	\mathbb{R}^{n_1\times r_1}_{+}, \mat{B} \in 	\mathbb{R}^{n_2 \times r_2}_{+}, \mat{C} \in 	\mathbb{R}^{n_3 \times r_3}_{+} \\
  \ten{G}^1 \in \mathbb{R}^{r_2\times r_3\times m}_{+}, \ten{G}^2 \in \mathbb{R}^{r_1 \times r_3 \times m}_{+}, \ten{G}^3 \in \mathbb{R}^{r_1 \times r_2 \times m}_{+}.
\end{multline*}
We enforce non-negativity constraints on the matrices $\mat{A}, \mat{B}, \mat{C}$ and the core tensors $\ten{G}^i$ ($i \in \{1, 2, 3\}$). Non-negativity is essential for learning interpretable latent factors \cite{Murphy2012}.

Each slice of the core tensor $\ten{G}^3$ corresponds to one of the $m$ relations. Each cell in a slice corresponds to an induced schema in terms of the latent factors from matrices $\mat{A}$ and $\mat{B}$. In other words, $\ten{G}^3_{i,j,k}$ is an induced binary schema for relation $k$ involving induced categories represented by columns $\mat{A}_{i}$ and $\mat{B}_{j}$. Cells in $\ten{G}^1$ and $\ten{G}^2$ may be interpreted accordingly. 

We derive non-negative multiplicative updates for $\mat{A}, \mat{B}$ and  $\mat{C}$  following the NMF updating rules given in \cite{Lee00algorithmsfor}. For the update of $\mat{A}$, we consider the mode-1 matricization of first and the second term in \refeqn{eqn:jtf_eqn}
along with the regularizer. 

 \begin{multline*}
  \mat{A}  \gets \mat{A} * \frac{\ten{X}^3_{(1)}\mathcal{G}_{B_A}^{\top} + \ten{X}^2_{(1)}\mathcal{G}_{C_A}^{\top}}{\mat{A}[\mathcal{G}_{B_A}\mathcal{G}_{B_A}^{\top} + \mathcal{G}_{C_A}\mathcal{G}_{C_A}^{\top}]+\lambda_a \mat{A}},
 \end{multline*}
 where,
 \[
  \mathcal{G}_{B_A} = (\ten{G}^3\times_2 \mat{B})_{(1)}, ~~ \mathcal{G}_{C_A} = (\ten{G}^2\times_2 \mat{C})_{(1)}.
 \]
 
In order to estimate $\mat{B}$, we consider mode-2 matricization of first term and mode-1 matricization of third term in \refeqn{eqn:jtf_eqn}, along with the regularization term. We get the following update rule for $\mat{B}$

 \begin{multline*}
  \mat{B} \gets \mat{B} * \frac{\ten{X}^3_{(2)}\mathcal{G}_{A_B}^{\top} + \ten{X}^1_{(1)}\mathcal{G}_{C_B}^{\top}}{\mat{B}[\mathcal{G}_{A_B}\mathcal{G}_{A_B}^{\top} + \mathcal{G}_{C_B}\mathcal{G}_{C_B}^{\top}]+\lambda_b \mat{B}},
 \end{multline*}
 
 where,

 \[
  \mathcal{G}_{A_B} = (\ten{G}^3\times_1 \mat{A})_{(2)}, ~~ \mathcal{G}_{C_B} = (\ten{G}^1\times_2 \mat{C})_{(1)}.
 \]
 
  For updating $\mat{C}$, we consider mode-2 matricization of second and third terms in \refeqn{eqn:jtf_eqn} along with the regularization term, and we get

 \begin{multline*}
  \mat{C} \gets \mat{C} * \frac{\ten{X}^3_{(2)}\mathcal{G}_{B_C}^{\top}+ \ten{X}^2_{(2)}\mathcal{G}_{A_C}^{\top}}{\mat{C}[\mathcal{G}_{A_C}\mathcal{G}_{A_C}^{\top} + \mathcal{G}_{B_C}\mathcal{G}_{B_C}^{\top}]+\lambda_c \mat{C}},
 \end{multline*}

 where,

 \[
  \mathcal{G}_{A_C} = (\ten{G}^3\times_1 \mat{B})_{(2)}, ~~ \mathcal{G}_{B_C} = (\ten{G}^2\times_1 \mat{A})_{(2)}.
 \]

Finally, we update the three core tensors  in \refeqn{eqn:jtf_eqn} following \cite{KimC07} as follows,

\[
 \ten{G}^1 \gets \ten{G}^1 * \frac{\ten{X}^1 \times_1 \mat{B}^{\top} \times_2 \mat{C}^{\top}}{\ten{G}^1 \times_1 \mat{B}^{\top}\mat{B} \times_2 \mat{C}^{\top}\mat{C}} ,  
\]
\[
 \ten{G}^2 \gets \ten{G}^2 * \frac{\ten{X}^2 \times_1 \mat{A}^{\top} \times_2 \mat{C}^{\top}}{\ten{G}^2 \times_1 \mat{A}^{\top}\mat{A} \times_2 \mat{C}^{\top}\mat{C}} ,  
\]
\[
 \ten{G}^3 \gets \ten{G}^3 * \frac{\ten{X}^3 \times_1 \mat{A}^{\top} \times_2 \mat{B}^{\top}}{\ten{G}^3 \times_1 \mat{A}^{\top}\mat{A} \times_2 \mat{B}^{\top}\mat{B}} .
\]

In all the above updates,  $\frac{\mat{P}}{\mat{Q}}$ represents element-wise division and $\mat{I}$ is the identity matrix.

{\bf Initialization}: For initializing the component matrices $\mat{A}, \mat{B}$, and $\mat{C}$, we first perform a non-negative Tucker2 Decomposition of the individual input tensors  $\ten{X}^1, \ten{X}^2,$ and $\ten{X}^3$.  Then compute the average of component matrices obtained from each individual decomposition for initialization. We initialize the core tensors $\ten{G}^1, \ten{G}^2,$ and $\ten{G}^3$  with the core tensors
obtained from the individual decompositions. 

\subsubsection{Step 2: Binary to Higher-Order Schema Induction}
\label{sec:schema}

In this section, we describe how a higher-order schema is constructed from the factorization described in the previous sub-section. Each relation $k$ has three representations  given by the slices $\ten{G}^1_k$, $\ten{G}^2_k$ and $\ten{G}^3_k$ from each core tensor. We need a principled
way to produce a joint schema from these representations. For a relation, we select top-$n$ indices $(i,j)$ with highest values from each matrix. The indices $i$ and $j$ from $\ten{G}^3_k$ correspond to column numbers of $\mat{A}$ and $\mat{B}$
respectively, indices from $\ten{G}^2_k$ correspond to columns from $\mat{A}$ and $\mat{C}$ and columns from $\ten{G}^1_k$ correspond to columns from $\mat{B}$ and $\mat{C}$.

We construct a tri-partite graph with the column numbers from each of the 
component matrices $\mat{A}$, $\mat{B}$ and $\mat{C}$ as the vertices belonging to independent sets, the top-$n$ indices selected are the edges between these vertices. From this tri-partite graph, we find all the triangles which will give schema
with three arguments for a relation, illustrated in \reffig{fig:schemata}. We find higher order schemata, i.e., schemata with more than three arguments by merging two third order schemata with same column number from $\mat{A}$ and $\mat{B}$. For example, if we find two schemata  
$(\mat{A}_2, \mat{B}_4, \mat{C}_{10})$ and $(\mat{A}_2, \mat{B}_4, \mat{C}_8)$ then we  merge these two to give $(\mat{A}_2, \mat{B}_4, \mat{C}_{10}, \mat{C}_8)$ as a higher order schema. This can be continued
further for even higher order schemata. This process may be thought of as finding a \textbf{constrained clique} over the tri-partite graph. Here the constraint is that in the maximal clique, there can only be one edge between sets corresponding
to columns of $\mat{A}$ and columns of $\mat{B}$.

The procedure above is inspired by \cite{McDonald:2005:SAC:1219840.1219901}. However, we note that \cite{McDonald:2005:SAC:1219840.1219901} solved a different problem, viz., n-ary relation instance extraction, while our focus is on inducing  schemata. 
Though we  discuss the case of back-off from 4-order to 3-order, ideas presented above can be extended for even higher orders depending on the sparsity of the tensors.

\section{Experiments}
\label{sec:exp}

\begin{table*}[t]
\begin{center}
   \begin{small}

 \begin{tabular}{|p{2.5cm}|c|c|c|}
 \hline
 \textbf{Dataset} & \textbf{$\mathcal{X}^1 shape$} &\textbf{$\mathcal{X}^2 shape$}&\textbf{$\mathcal{X}^3 shape$}  \\
 \hline
\hline	  
	   Shootings &   $3365\times 1295 \times 50$  & $2569\times 1295 \times 50$  & $2569 \times 3365 \times 50$ \\
		NYT Sports & $57,820 \times 20,356 \times 50$   & $49,659 \times 20,356 \times 50$  &  $49,659 \times 57,820 \times 50$\\	 
	 MUC & $ 2825 \times 962 \times 50$ & $2555 \times 962 \times 50$& $2555 \times 2825 \times 50 $\\
 \hline
\end{tabular}
\caption{\label{tbl:datasets}Details of dimensions of tensors constructed for each dataset used in the experiments.}
\end{small}
\end{center}
\end{table*}

\begin{table}[t]
 \centering
 \small
 \begin{tabular}{|l|c|c|}
 \hline
   Dataset & $(r_1, r_2, r_3)$ & $(\lambda_a, \lambda_b, \lambda_c)$ \\
   \hline
	\hline
   Shootings & (10, 20,15) & (0.3, 0.1, 0.7) \\
   NYT Sports & (20, 15, 15)&  (0.9, 0.5, 0.7)\\   
   MUC & (15, 12, 12) &  (0.7, 0.7, 0.4)\\
   \hline
 \end{tabular}
\caption{\label{tbl:hyperparameters} Details of hyper-parameters set for different datasets.}
\end{table}

\begin{table*}[tb]
\begin{scriptsize}
 \centering
 \begin{tabular}{|p{3cm}|p{7cm}|p{1.2cm}|p{2.5cm}|}
 \hline
 \multicolumn{1}{|c|}{Relation Schema} & \multicolumn{1}{|c|}{NPs from the induced categories} & \multicolumn{1}{|c|}{Evaluator Judgment} & \multicolumn{1}{|c|}{(Human) Suggested Label}\\
  \hline
 \multicolumn{4}{|c|}{Shootings} \\
 \hline
\multirow{3}{*}{\textit{leave}$\langle A_6, B_0,C_7 \rangle$}  & $A_6$: \textit{shooting, shooting incident, double shooting} & \multirow{3}{*}{valid} & $<$ shooting $>$\\
   & $B_0$: \textit{one person, two people, three people} & & $<$ people $>$ \\
   & $C_7$: \textit{dead, injured, on edge} & & $<$injured $>$ \\
  \hline
  \multirow{4}{*}{\textit{identify}$\langle A_{1},B_{1},C_{5}, C_{6} \rangle$}  & $A_{1}$: \textit{police, officers, huntsville police}  & \multirow{4}{*}{valid} & $<$ police $>$\\ 
   & $B_{1}$: \textit{man, victims, four victims} & & $<$ victim(s)$>$ \\
   & $C_{5}$: \textit{sunday, shooting staurday, wednesday afternoon} & & $<$day/time $>$ \\
   & $C_{6}$: \textit{apartment, bedroom, building in the neighborhood} & & $<$place $>$\\
   \hline
   \multirow{3}{*}{\textit{shoot}$\langle A_{7},B_{6}, C_{1} \rangle$}  & $A_{7}$: \textit{gunman, shooter, smith} & \multirow{3}{*}{valid} & {$<$ perpetrator $>$}\\ 
   & $B_{6}$: \textit{freeman, slain woman, victims} & & $<$victim $>$\\  
   & $C_{1}$: \textit{friday, friday night, early monday morning} & & $<$ time$>$\\ 
   \cline{2-4}
   \multirow{3}{*}{\textit{shoot}$\langle A_{4},B_{2}, C_{13} \rangle$}  & $A_{4}$: \textit{$<$num$>$-year-old man, $<$num$>$-year-old george reavis, $<$num$>$-year-old brockton man} & \multirow{3}{*}{valid} & $<$ victim$>$\\ 
   & $B_{2}$: \textit{in the leg, in the head, in the neck} & & $<$ body part$>$\\  
   & $C_{13}$: \textit{in macon, in chicago, in an alley} & & $<$ location $>$\\ 
   \hline   
   \multirow{3}{*}{\textit{say}$\langle A_{1},B_{1}, C_{5} \rangle$}  & $A_{1}$: \textit{police, officers, huntsville police} & \multirow{3}{*}{invalid} & \multirow{3}{*}{--}\\ 
   & $B_{1}$: \textit{man, victims, four victims} & & \\  
   & $C_{5}$: \textit{sunday, shooting staurday, wednesday afternoon} & & \\ 
  \hline  
  \multicolumn{4}{|c|}{NYT sports} \\
  \hline
  \multirow{3}{*}{\textit{spend}$\langle A_{0},B_{16},C_{3} \rangle$}  & $A_{0}$: \textit{yankees, mets, jets }  & \multirow{3}{*}{valid}  &$<$ team $>$ \\ 
   & $B_{14}$: \textit{\$ $<$num$>$ million, \$ $<$num$>$, \$ $<$num$>$ billion } & &$<$ money $>$\\ 
   & $C_{3}$: \textit{$<$num$>$, year, last season} & & $<$ year $>$\\ 
   \hline
  \multirow{3}{*}{\textit{win}$\langle A_{2},B_{10},C_{3} \rangle$}  & $A_2$: \textit{red sox, team, yankees} & \multirow{3}{*}{valid} & $<$ team $>$ \\ 
   & $B_{10}$: \textit{world series, title, world cup} & & $<$ championship $>$ \\ 
   & $C_{3}$: \textit{$<$num$>$, year, last season} & & $<$ year $>$ \\
   \hline
  \multirow{3}{*}{\textit{get}$\langle A_{4},B_{4},C_{1} \rangle$}  & $A_{4}$: \textit{umpire, mike cameron, andre agassi} & \multirow{3}{*}{invalid}& \multirow{3}{*}{--}\\ 
   & $B_{4}$: \textit{ball, lives, grounder} & &\\
   & $C_{1}$: \textit{back, forward, $<$num$>$-yard line} & &\\
   \hline          
  \multicolumn{4}{|c|}{MUC} \\
  \hline
  \multirow{3}{*}{\textit{tell}$\langle A_{7},B_{2},C_{0} \rangle$}  & $A_{7}$: \textit{medardo gomez, jose azcona, gregorio roza chavez}  & \multirow{3}{*}{valid}  &$<$ politician $>$ \\ 
   & $B_{2}$: \textit{media, reporters, newsmen} & & $<$media $>$\\ 
   & $C_{0}$: \textit{today, at $<$num$>$, tonight} & & $<$ day/time $>$\\ 
   \hline  
  \multirow{3}{*}{\textit{occur}$\langle A_{9},B_{5},C_{10} \rangle$}  & $A_9$: \textit{bomb, blast, explosion} & \multirow{3}{*}{valid} & $<$ bombing $>$ \\ 
   & $B_{5}$: \textit{near san salvador, here in madrid, in the same office} & & $<$ place $>$ \\ 
   & $C_{10}$: \textit{at $<$num$>$, this time, simultaneously} & & $<$ time $>$ \\
   \hline  
  \multirow{3}{*}{\textit{suffer}$\langle A_{5},B_{4},C_{4})$}  & $A_{5}$: \textit{justice maria elena diaz, vargas escobar, judge sofia de roldan} & \multirow{3}{*}{invalid}& \multirow{3}{*}{--}\\ 
   & $B_{4}$: \textit{casualties , car bomb, grenade} & &\\
   & $C_{4}$: \textit{settlement of refugees, in san roman, now } & &\\
   \hline

   \hline
 
 \end{tabular}
 
 \caption{\label{tbl:predicate_examples}Examples of  schemata induced by \system. Please note that some of them are 3-ary while others are 4-ary.  For details about schema induction, please refer to \refsec{sec:model}.}
\end{scriptsize}
\end{table*}

\begin{table*}[!tb]
\centering
\begin{small}
 \begin{tabular}{|c|c|c|c|c|c|c|c|c|c|c|c|c|}
  \hline
  
   \multirow{2}{*}{}& \multicolumn{4}{|c|}{Shootings} &\multicolumn{4}{|c|}{NYT Sports} &\multicolumn{4}{|c|}{MUC}\\
  \cline{2-13}
  & E1 & E2 & E3 &Avg & E1 &E2 & E3& Avg & E1 & E2 &E3 &  Avg\\ \hline \hline
\HCB{} & 0.64& 0.70&0.64 & 0.66 & 0.42 &0.28  & 0.52 &0.46 & 0.64 & 0.58 & 0.52 & \textbf{0.58}\\ 
Chambers-13 & 0.32& 0.42 &0.28 & 0.34 & 0.08 &0.02 & 0.04& 0.07 & 0.28 & 0.34 & 0.30 &  0.30\\ 
\system{}  & 0.82&  0.78 & 0.68 & \textbf{0.76} & 0.86 & 0.6 & 0.64 &\textbf{0.70} & 0.58 & 0.38 & 0.48 & 0.48\\ \hline

 \end{tabular}
 \end{small}
\caption{\label{tbl:accuracy}Higher-order RSI accuracies of various methods on the three datasets. %\system{} compared to the baseline \cite{Chambers13} model. 
Induced schemata for each dataset and method are evaluated by three human evaluators, E1, E2, and E3. \system{} performs better than \HCB{} for Shootings and NYT Sports datasets. Even though \HCB{} achieves better accuracy on MUC dataset, it has several limitations, see \refsec{sec:exp} for more details.
Chambers-13 solves a slightly different problem called event schema induction, for more details about the comparison with Chambers-13 see \refsec{sec:chambers-13}.}
\end{table*}

In this section, we evaluate the performance of \system{} for the task of \problem{}. We also propose a baseline model for \problem{} called HardClust. \\
{\bf HardClust}: We propose a baseline model called the Hard Clustering Baseline (\HCB{}) for the task of higher order relation schema induction.
This model induces schemata by grouping per-relation NP arguments from OpenIE extractions. In other words, for each relation, all the Noun Phrases (NPs) in first argument form a cluster that represents the  subject of the relation, all the NPs in the second argument form a cluster that represents object and so on. 
Then from each cluster, the top most frequent NPs are chosen as the representative NPs for the argument type. We note that this method is only able to induce one schema per relation.

\textbf{Datasets:}  We run our experiments on three datasets. The first dataset (Shootings) is a collection of 1,335 documents constructed from a publicly available database of mass shootings in the United States. The second 
is New York Times Sports (NYT Sports) dataset which is a collection of 20,940 sports documents from the period 2005 and 2007. And the third dataset (MUC) is a set of 1300 Latin American newswire documents about terrorism events. 
After performing the processing steps described in \refsec{sec:method}, 
we obtained 357,914 unique OpenIE extractions from the NYT Sports dataset, 10,847 from Shootings dataset, and 8,318 from the MUC dataset. 
However, in order to properly analyze and evaluate the model, we consider only the 50 most frequent relations in the datasets and their corresponding OpenIE extractions.
This is done to avoid noisy OpenIE extractions to yield better data quality and to aid subsequent manual evaluation of the data.
We construct input tensors following the procedure described in \refsec{sec:model}.
Details on the dimensions of tensors obtained are given in \reftbl{tbl:datasets}.

\textbf{Model Selection}: In order to select appropriate \system{} parameters, we perform a grid search over the space of hyper-parameters, and select the set of hyper-parameters that give best Average FIT score ($\mathrm{AvgFIT}$).  % which is defined as follows:
\begin{multline*}
 \mathrm{AvgFIT}(\ten{G}^1, \ten{G}^2, \ten{G}^3, \mat{A}, \mat{B}, \mat{C}, \ten{X}^1, \ten{X}^2, \ten{X}^3) = \\
 \frac{1}{3} \{\mathrm{FIT}(\ten{X}^1, \ten{G}^1, \mat{B}, \mat{C}) + \mathrm{FIT}(\ten{X}^2, \ten{G}^2, \mat{A}, \mat{C})\\
 + \mathrm{FIT}(\ten{X}^3, \ten{G}^3, \mat{A}, \mat{B})\},
\end{multline*}
 where,
 \begin{multline*}
  \mathrm{FIT}(\ten{X}, \ten{G}, \mat{P}, \mat{Q}) = 1 - \frac{\norm{\ten{X} - \ten{G}\times_1 \mat{P} \times_2 \mat{Q}}_F }{\norm{\ten{X}}_F}.
 \end{multline*}

We perform a grid search for the rank parameters between 5 and 20, for the regularization weights we perform a grid search over 0 and 1. \reftbl{tbl:hyperparameters} provides the details of hyper-parameters set for different datasets.
\textbf{Evaluation Protocol}: For \system{}, we follow the protocol mentioned in \refsec{sec:schema} for constructing higher order schemata. For every relation, we consider top 5 binary schemata from the factorization of each tensor. 
We construct a tripartite graph, as explained in \refsec{sec:schema}, and mine constrained maximal cliques from the tripartite graphs for schemata. \reftbl{tbl:predicate_examples} provides some qualitative examples of higher-order schemata induced by \system{}. Accuracy of the schemata induced by the model is evaluated by human evaluators. In our experiments, we use  human judgments from three evaluators. 
For every relation, the first and second columns given in \reftbl{tbl:predicate_examples} are presented to the evaluators and they are asked to validate the schema.
We present top 50 schemata based on the score of the constrained maximal clique induced by \system{} to the evaluators.
This evaluation protocol was also used in \cite{kblda:movshovitzattias-wcohen:2015:ACL} for evaluating ontology induction. All evaluations were blind, i.e., the evaluators were not aware of the model they were evaluating. 

{\bf Difficulty with Computing Recall}: Even though recall is a desirable measure, due to the lack of  availability of gold higher-order schema annotated corpus, it is not possible to compute recall. Although the MUC dataset has gold annotations for some predefined list of events, it does not have annotations for the relations.

 Experimental results comparing performance of various models for the task of \problem{} are given in \reftbl{tbl:accuracy}. We present evaluation results from three evaluators represented as E1, E2 and E3. 
As can be observed from \reftbl{tbl:accuracy}, \system{} achieves better results than \HCB{} for the Shootings and NYT Sports datasets, however \HCB{} achieves better results for the MUC dataset. Percentage agreement of the evaluators for \system{} is 72\%, 70\% and 60\% for Shootings, NYT Sports and MUC datasets respectively.

\textbf{\HCB{} Limitations}: Even though \HCB{}  gives better induction for MUC corpus, this approach has some serious drawbacks. \HCB{} can only induce one schema per relation. This is a restrictive constraint as multiple senses can exist for a relation.
For example, consider the schemata induced for the relation \textit{shoot} as shown in \reftbl{tbl:predicate_examples}. \system{} induces two senses for the relation, but \HCB{} can induce only one schema.
For a set of 4-tuples, \HCB{} can only induce ternary schemata; the dimensionality of the schemata cannot be varied. Since the latent factors induced by \HCB{} are entirely based on  frequency, the latent categories induced by \HCB{} are dominated by only a fixed set of noun phrases.
For example, in NYT Sports dataset, subject category induced by \HCB{} for all the relations is $\langle${\it team, yankees, mets}$\rangle$. In addition to inducing only one schema per relation, most of the times \HCB{} only induces a fixed set of categories. Whereas for \system{}, the number of categories depends on the rank of factorization, which is a user provided parameter, thus providing more flexibility to choose the latent categories. 
\subsection{Using Event Schema Induction for \problem{}}
\label{sec:chambers-13}

Event schema induction is defined as the task of learning high-level representations of  events, like a tournament,  and their entity roles, like winning-player etc,   from unlabeled text. Even though the main focus of event schema induction is to induce the important roles of the events,  as a side result most of the algorithms also provide schemata for the relations. 
In this section, we investigate the effectiveness of these schemata compared to the ones induced by \system{}. 

Event schemata are represented as a set of {\it (Actor, Rel, Actor)} triples in \cite{balasubramanian-EtAl:2013:EMNLP}. {\it Actors } represent groups of noun phrases and {\it Rel}s represent relations. From this style of representation, however, the n-ary schemata for relations cannot be induced. 
Event schemata generated in \cite{Weber2018} are similar to that in \cite{balasubramanian-EtAl:2013:EMNLP}.
Event schema induction algorithm proposed in \cite{nguyen-EtAl:2015:ACL-IJCNLP1} doesn't induce schemata for relations, but rather induces the roles for the events. For this investigation we experiment with the following algorithm.

{\bf Chambers-13} \cite{Chambers13}:  This model learns event templates from text documents. Each event template provides a distribution over slots, where slots are clusters of NPs. Each event template also provides a cluster of relations, which is most likely to appear in the context of the aforementioned slots. We evaluate the schemata of these relation clusters.

As can be observed from \reftbl{tbl:accuracy}, the proposed \system{} performs much better than Chambers-13. \HCB{} also performs better than Chambers-13 on all the datasets. From this analysis we infer that there is a need for 
algorithms which induce higher-order schemata for relations, a gap we fill in this paper. Please note that the experimental results provided in \cite{Chambers13} for MUC dataset are for the task of event schema induction, but in this work we evaluate the relation schemata. Hence the results in \cite{Chambers13}  and results in this paper are not comparable. Example schemata induced by \system{} and (Chambers-13) are provided as part of the supplementary material.

\section{Conclusion}
\label{sec:conclusion}

Higher order Relation Schema Induction (\problem{}) is an important first step towards building domain-specific Knowledge Graphs (KGs). In this paper, we proposed \system{}, a  tensor factorization-based method for higher-order RSI. To the best of our knowledge, this is the first attempt at inducing higher-order (n-ary) schemata for relations from unlabeled text. Rather than factorizing a severely sparse higher-order tensor directly, \system{} performs  \emph{back-off} and jointly factorizes multiple lower-order tensors derived out of the higher-order tensor. In the second step, \system{} solves a constrained clique problem to induce schemata out of multiple binary schemata. 
We are hopeful that the backoff-based factorization idea exploited in \system{} will be useful in other sparse factorization settings.

\pagebreak
\section*{Acknowledgment}

We thank the anonymous reviewers for their insightful comments and suggestions. This research has been supported in part by the Ministry of Human Resource Development (Government of India), Accenture, and Google.

\bibliography{main}
\bibliographystyle{styles/acl_natbib}

\end{document}